\documentclass{easychair}

\usepackage{amsmath}
\usepackage{enumitem}
\usepackage{tikz}
\usepackage[]{algorithm2e}
\usepackage{hyphenat}
\usetikzlibrary{matrix}

\usepackage{doc}

\newcommand{\tryhard}{\texttt{try\_hard}}
\newcommand{\Comment}[1]{}
\title{Towards Smart Proof Search for Isabelle}

\author{
Yutaka Nagashima%\thanks{Designed and implemented the class style}
}

\institute{
  Data61, CSIRO,
  Sydney, New South Wales, Australia\\
  \email{first\_name.last\_name@data61.csiro.au}
 }

\authorrunning{Yutaka Nagashima}

\titlerunning{Towards Smart Proof Search for Isabelle}
\begin{document}

\maketitle

\begin{abstract}
Despite the recent progress in automatic theorem provers,
proof engineers are still suffering from the lack of powerful proof automation.
In this position paper
we first report our proof strategy language based on a meta-tool approach.
%as well as the advantages and disadvantages.
Then, we propose an AI-based approach to drastically improve proof automation for Isabelle,
while identifying three major challenges we plan to address for this objective.
\end{abstract}

% The table of contents below is added for your convenience. Please do not use
% the table of contents if you are preparing your paper for publication in the
% EPiC or Kalpa series

%\setcounter{tocdepth}{2}
%{\small
%\tableofcontents}

%\section{To mention}
%
%Processing in EasyChair - number of pages.
%
%Examples of how EasyChair processes papers. Caveats (replacement of EC
%class, errors).

\pagestyle{empty}

%------------------------------------------------------------------------------
\section{PSL and Meta-Tool Approach}

In the last decade, we have seen the successful application of automated theorem provers
to assist interactive theorem proving\Comment{Cite Sledgehammer and Holy Hammer}.
Despite the popularity of these so-called ``hammer-style'' tools,
their performance is still suffering from the gap between underlying logics \cite{DBLP:journals/jfrea/BlanchetteKPU16}.

To circumvent this problem,
we introduced a proof strategy language, PSL \cite{DBLP:journals/corr/NagashimaK16}, to 
Isabelle/HOL \cite{DBLP:books/sp/NipkowPW02},
taking a meta-tool approach.
A proof strategy is an abstract description of how to attack proof obligations.
Users write strategies in PSL based on their intuitions on a conjecture.
Using a strategy, PSL's runtime system generates many invocations of 
Isabelle's native proof tools, called \textit{tactics}.
The runtime tries to find out the appropriate combination of
tactics for each goal by trial-and-error.
This way, PSL reduces the domain specific procedure of interactive theorem proving
to the well-known dynamic tree search problem.
The meta-tool language approach brought the following advantages: 
\begin{itemize}[noitemsep]
\item Domain specific procedures can be added as new sub-tools.
\item Sub-tools (including Sledgehammer) can be improved independently of PSL.
\item Generated efficient-proof scripts are native Isabelle proofs.
\end{itemize}

We provided a default strategy, \tryhard{}.
Our evaluation shows that
PSL based on \tryhard{} significantly outperforms Sledgehammer for many use cases \cite{DBLP:journals/corr/NagashimaK16}.
However, PSL's proof search procedure is still mostly brute-force.
For example, when PSL generates many tactics,
even though each of these is tailored for the proof goal utilizing the runtime information,
it is still the statically written strategy, \tryhard{}, that decides:
\begin{itemize}[noitemsep]
  \item what kind of tactics to generate,
  \item how to combine them, and
  \item in which order to apply generated tactics.
\end{itemize}

%------------------------------------------------------------------------------

\section{Meta-Tool Based Smart Proof Search}
On a fundamental level,
this lack of flexibility stems from the procedural nature of Isabelle's tactic language,
which describes how to prove conjectures in a step-by-step manner
rather than what to do with conjectures.
However, a conventional declarative language would not be a good solution either,
since in many cases we cannot determine 
what to do with a given proof goal with certainty.
Our guess may or may not be right:
given a proof goal, 
we cannot tell which tactics to use until we complete the goal.
%Furthermore, there are usually multiple effective ways to attack a proof obligation;
%however, a language to describe what to do deterministically
%cannot directly express this situation.

\paragraph{Probabilistic Declarative Proof Strategy Language.}
We propose the development of a probabilistic declarative proof strategy language (PDPSL),
which allows for the description of feasible tactics 
for \textit{arbitrary} proof obligation.
For example, when we apply 
a proof strategy, \verb|str|, written in PDPSL,
to a proof search graph, \verb|pgraph|,
PDPSL's runtime: 
\begin{itemize}[noitemsep]
\item [(1)] picks up the most promising node in \verb|pgraph|,
\item [(2)] applies the most promising tactic that has not been applied to that node yet, and
\item [(3)] adds the resultant node to \verb|pgraph|, 
connecting them with a labelled edge representing the tactic application 
and weight indicating how promising that tactic application is. % against the picked node.
\end{itemize}
The runtime keeps applying this procedure until 
it reaches a solved state, in which the proof is complete.
This way, PDPSL executes a best-first search at runtime.

The major challenge is the design of PDPSL.
%Ideally, PDPSL should be able to describe the engineers' intuitions on proof development 
A strategy written in PDPSL is applied repeatedly during a best-first search
against emerging proof obligations,
and we cannot predict 
how these intermediate goals look like prior to the search;
therefore, weighting of tactics against concrete proof goals is not good enough.
PDPSL should only describe the meta-information on proof goals
and information in Isabelle's standard library.
Such meta-information includes:

\begin{itemize}[noitemsep]
  \item Which ITP mechanism was used to define constants that appear in the proof obligation?
%  \item Does Isabelle already know induction schemes that involve these constants?
  \item Does the conjecture involve recursive functions?
%  \item where in abstract syntax tree each variable appears.
\end{itemize}

%Presumably, PDPSL will be based on probabilistic programming model
%but specialised for ITP.
This may sound too restrictive;
however, when proof engineers use ITPs,
we often reason on the meta level.
For instance, Isabelle's tutorial \cite{DBLP:books/sp/NipkowPW02} introduces the following heuristics:
\begin{itemize}[noitemsep]
  \item []``\textit{Theorems about recursive functions are proved by induction.}''
  \item []``\textit{The right-hand side of an equation should (in some sense) be simpler
than the left-hand side.}''
\end{itemize}
These are independent of user-defined formalisation.

\paragraph{Posterior Proof Attempt Evaluation.}
Evaluating a proof goal to write promising tactics is only half the story of 
typical interactive proof development.
The other half is about evaluating the results of tactic applications.
We presume that posterior evaluations on tactic application will improve proof search
when combined with prior expectations on tactics described in PDPSL.
The main challenge is to find a useful measure by:
\begin{itemize}[noitemsep]
  \item discovering (possibly multiple) useful measures that indicate progress in proof search, and
  \item investigating how to integrate these measures into the best-first search of PDPSL.
\end{itemize}

Of course it is not possible to come up with a heuristic 
that is guaranteed to work for all kinds of proof obligations.
But when focusing on a particular problem domain such as systems verification,
it is plausible to find a useful measure 
to narrow the search space at runtime.

%Putting together priori-and-posteriori heuristics, 
%Algorithm \ref{alg:bfs} roughly shows my idea of how the proposed deep smart proof search should work.

%\begin{algorithm} %[H]
% \KwData{proof obligation ($PO$) and a strategy ($str$) in PDPSL}
% \KwResult{efficient-proof script}
% enqueue $PO$ into a priority queue, $PQ$.\;
% \While{The solved state is not found.}{
%  $n0$ $\leftarrow$ pop $PQ$\;
%  $tac$ $\leftarrow$ create the most promising tactic not applied to $n$\;
%  $n1$ $\leftarrow$ apply $tac$ to $n$\;
%  $n2$ $\leftarrow$ evaluate $n1$'s expected distance to the solved state\;
%  $n0$ $\leftarrow$ tag $n0$ with $tac$ to indicate $tac$ is already tried.\;
%  enqueue $n0$\;
%  enqueue $n2$\;
% }
% print the path to the solved node as the fast proof script.
% \caption{Deep Smart Proof Search} \label{alg:bfs}
%\end{algorithm}

\paragraph{Reinforcement Learning of PDPSL using Large Proof Corpora.}
PDPSL is primarily for engineers to encode their intuitions.
But at the next step, we plan to improve this hand-written heuristics 
using reinforcement learning on the existing proof corpora.
The Isabelle community has a repository of formal proofs, 
called the Archive of Formal Proofs (AFP) \cite{AFP}.
As of 2015, the size of the AFP is larger than one million lines of code.
%Ignoring the definitions and lemma statements, we still have more than
%590,000 lines of carefully hand written proof scripts \cite{mining}.
Additionally,
the seL4 project open-sourced roughly 400,000 lines of Isabelle proof script \cite{Klein_14_2}.
We speculate that these large proof corpora will work as the basis 
for reinforcement learning of strategies written in PDPSL.
The lack of concrete information on proof obligations can be an advantage at this stage:
since PDPSL cannot describe concrete proof obligations,
it is less likely to cause over-fitting.

%\paragraph{Large Scale Parallelism.}
%I plan to exploit parallelism to make proof search as powerful as possible.
%Though parallelism is often a hard problem, 
%it will be probably not the case for PDPSL,
%because proof search is mostly computation-intensive procedure 
%which is sometimes described as \textit{embarrassingly parallel}.
%I will probably implement the above mentioned best-first search using a priority queue.
%I expect that simple locking and releasing of the queue to be good enough to gain performance gain out of large scale parallelism, 
%since each step of tactic application tends to take significant length of time.
%If this is not the case, I have to resort to more involved data-structure, 
%such as \verb|SplayList|.

%\section{Comparison and Related Work}
%\begin{itemize}
%  \item Matryoshka and Sledgehammer.
%  \item Joseph Urban.
%  \item Cezery .
%  \item Importance to integrate humans' expertise inside automation.
%\end{itemize}

\label{sect:bib}
\bibliographystyle{plain}
\bibliography{easychair}

%------------------------------------------------------------------------------
% Index
%\printindex

%------------------------------------------------------------------------------
\end{document}